\documentclass{article}

\PassOptionsToPackage{numbers, compress}{natbib}

\usepackage[final]{nips_2016}
\usepackage{nips_2016}
\usepackage[utf8]{inputenc} 
\usepackage[T1]{fontenc}    
\usepackage{hyperref}       
\usepackage{url}            
\usepackage{booktabs}       
\usepackage{amsfonts}       
\usepackage{nicefrac}       
\usepackage{microtype}      
\usepackage{graphicx}		
\usepackage{mathtools}		
\usepackage{textcomp}
\usepackage{multicol}
\usepackage{color}

\title{Hierarchical Object Detection \\with Deep Reinforcement Learning}

\author{
  Míriam Bellver Bueno\\
  Barcelona Supercomputing Center (BSC)\\
  Barcelona, Catalonia/Spain\\
  \texttt{miriam.bellver@bsc.es} \\
  \And
  Xavier Giró-i-Nieto \\
  Image Processing Group \\
  Universitat Politècnica de Catalunya (UPC) \\
  Barcelona, Catalonia/Spain  \\
  \texttt{xavier.giro@upc.edu} \\
  \AND
  Ferran Marqués \\
  Image Processing Group \\
  Universitat Politècnica de Catalunya (UPC)\\
  Barcelona, Catalonia/Spain  \\
  \texttt{ferran.marques@upc.edu} \\
  \And
  Jordi Torres  \\
  Barcelona Supercomputing Center (BSC)\\
  Universitat Politècnica de Catalunya\\
  Barcelona, Catalonia/Spain\\
  \texttt{jordi.torres@bsc.es} \\
}

\begin{document}

\maketitle

\begin{abstract}
 We present a method for performing hierarchical object detection in images guided by a deep reinforcement learning agent. The key idea is to focus on those parts of the image that contain richer information and zoom on them. We train an intelligent agent that, given an image window, is capable of deciding where to focus the attention among five different predefined region candidates (smaller windows). This procedure is iterated providing a hierarchical image analysis.We compare two different candidate proposal strategies to guide the object search: with and without overlap. Moreover, our work compares two different strategies to extract features from a convolutional neural network for each region proposal: a first one that computes new feature maps for each region proposal, and a second one that computes the feature maps for the whole image to later generate crops for each region proposal. Experiments indicate better results for the overlapping candidate proposal strategy and a loss of performance for the cropped image features due to the loss of spatial resolution. We argue that, while this loss seems unavoidable when working with large amounts of object candidates, the much more reduced amount of region proposals generated by our reinforcement learning agent allows considering to extract features for each location without sharing convolutional computation among regions. Source code and models are available at \url{https://imatge-upc.github.io/detection-2016-nipsws/}.

\end{abstract}

\section{Introduction}

When we humans look at an image, we always perform a sequential extraction of information in order to understand its content. First, we fix our gaze to the most salient part of the image and, from the extracted information, we guide our look towards another image point, until we have analyzed all its relevant information. This is our natural and instinctive behavior to gather information from our surroundings. 

Traditionally in computer vision, images have been analyzed at the local scale following a sliding window scanning, often at different scales. This approach analyses the different image parts independently, without relating them. Just by introducing a hierarchical representation of the image, we can more easily exploit the relationship between regions. We propose to use a top-down scanning which firstly takes a global view of the image to sequentially focus on the local parts that contain the relevant information (e.g. objects or faces). 

Our algorithm is based on an intelligent agent trained by reinforcement learning that is capable of making decisions to detect an object in a still image, similarly to \cite{caicedo2015active}. The agent first analyzes the whole image, and decides in which region of the image to focus among a set of predefined ones. Inspired by \cite{lu2015adaptive}, our agent can top-down explore a set of five different predefined region candidates: four regions representing the four quadrants plus a central region. Two different strategies have been studied: proposing overlapping or non-overlapping candidates. The agent stops its search when it finds an object. Reinforcement learning is useful for our task because there is no single way of completing it. The agent can explore the hierarchical representation in different ways and still achieve its goal. Then, instead of programming every step that the agent should do, we train it so that it makes decisions under uncertainty to reach its target.
Notice that the final goal of object detection is to define a bounding box around the object and that, in our work, these bounding boxes are limited to the predefined regions in the hierarchy.

Most state of the art solutions for object detection analyze large amounts of region proposals. These algorithms need to leverage the bottleneck of describing all these proposals by reusing convolutional feature maps of the whole image. In our work though, as the reinforcement learning agent and the hierarchy allow us to analyze a very reduced number of regions, we can feed each region visited by the agent through a convolutional network to extract its features, allowing us to work with region representations of higher spatial resolution, which are also more informative than those cropped from a feature map of the whole image. To study this trade-off, we have trained and compared two different models based on each of these two principles: the \textit{Image-Zooms} model, which extracts descriptors at each region, and the \textit{Pool45-Crops} model, which reuses feature maps for different regions of the same image.

The first contribution of our work is the introduction of a hierarchical representation to top-down (zoom-in) guide our agent through the image. We explore how the design of the hierarchy affects the detection performance and the amount of visited regions. The second contribution is the study between extracting features for each region instead of reusing feature maps for several locations. We show the gain of the region-specific features for our scheme, and argue that the computational overhead is minor thanks to the very reduced amount of regions considered by the agent.

\section{Related Work}
\label{gen_inst}

Reinforcement learning is a powerful tool that has been used in a wide range of applications. The most impressive results are those from DeepMind \citep{mnih2015human}, who have been able to train an agent that plays Atari 2600 video games by observing only their screen pixels, achieving even superhuman performance. Also they trained a computer that won the Go competition to a professional player for the first time \citep{silver2016mastering}. More specifically to traditional computer vision tasks, reinforcement learning has been applied to learn spatial glimpse policies for image classification \citep{mnih2014recurrent,ba2014multiple}, for captioning \citep{xu2015show} or for activity recognition \citep{yeung2015end}. It has also been applied for object detection in images \citep{caicedo2015active}, casting a Markov Decision Process, as our approach does.

The traditional solutions for object detection are based on region proposals, such as Selective Search \citep{van2011segmentation}, CPMC \citep{carreira2012cpmc} or MCG \citep{Pont-Tuset2016} and other methods based on sliding windows such as EdgeBoxes \citep{zitnick2014edge}. The extraction of such proposals was independent of the classifier that would score and select which regions compose the final detection. These methods are computationally expensive because rely on a large number of object proposals. Then the first trends based on Convolutional Neural Networks appeared, such as Fast R-CNN  \citep{girshick2015fast}, that already studied how to share convolutional computation among locations, as they identified that the extraction of features for the hypothesized objects was the bottleneck for object detection.

More recent proposals such as Faster R-CNN  \citep{ren2015faster} have achieved efficient and fast object detection by obtaining cost-free region proposals sharing full-image convolutional features with the detection network. Directly predicting bounding boxes from an image is a difficult task, and for this reason, approaches such as Faster R-CNN rely on a number of reference boxes called anchors, that facilitate the task of predicting accurate bounding boxes by regressing these initial reference boxes. One key of our approach is the refinement of bounding box predictions through the different actions selected by the reinforcement learning agent. Besides Faster R-CNN or other approaches such as YoLo \citep{redmon2015you} or MultiBox  \citep{erhan2014scalable} based on anchors, there are other works that are based on the refinement of predictions. \citet{yoo2015attentionnet} propose the AttentionNet. They cast an object detection problem as an iterative classification problem. AttentionNet predicts a number of weak directions pointing to the target object so that a final accurate bounding box is obtained. The state-of-the-art in object detection is the Single Shot MultiBox Detector (SSD) \citep{liu2015ssd}, which works with a number of default boxes of different aspect ratios and scales per each feature map location, and also adjusts them to a better match to the object shape.

Another approach that supports this idea is the Active Object Localization method proposed by \citet{caicedo2015active}. Their method trains an intelligent agent using deep reinforcement learning that is able to deform bounding boxes sequentially until they fit the target bounding box. Each action that the agent does to the bounding box can change its aspect ratio, scale or position. Our main difference to this approach is that we add a fixed hierarchical representation that forces a top-down search, so that each action zooms onto the predicted region of interest. 

How to benefit from super-resolution has also been studied by other works. In the paper of \citet{lu2015adaptive} a model is trained to determine if it is required to further divide the current observed region because there are still small objects on it, and in this case, each subregion is analyzed independently. Their approach could also be seen as hierarchical, but in this case they analyze each subregion when the zoom prediction is positive, whereas we just analyze the subregion selected by the reinforcement learning agent. On contrast to their proposal, we analyze fewer regions and then we can afford extracting high-quality descriptors for each of them, instead of sharing convolutional features.

\section{Hierarchical Object Detection Model}
\label{headings}

In this work we define the object detection problem as the sequential decision process of a goal-oriented agent interacting with a visual environment that is our image. At each time step the agent should decide in which region of the image to focus its attention so that it can find objects in a few steps. We cast the problem as a Markov Decision Process, that provides a framework to model decision making when outcomes are partly uncertain.

\subsection{MDP formulation}

In order to understand the models for the object detection task that we have developed, we first define how the Markov Decision Process is parameterized.

\paragraph{State}
\label{ssec:State}

The state is composed by the descriptor of the current region and a memory vector. The type of descriptor defines the two models we compare in our work: the \textit{Image-Zooms} model and the \textit{Pool45-Crops} model. These two variations are explained in detail in Section ~\ref{model_section}. The memory vector of the state captures the last 4 actions that the agent has already performed in the search for an object. As the agent is learning a refinement of a bounding box, a memory vector that encodes the state of this refinement procedure is useful to stabilize the search trajectories. We encode the past 4 actions in a one-shot vector. As there are 6 different actions presented in the following section, the memory vector has 24 dimensions. This type of memory vector was also used in \citep{caicedo2015active}.

\paragraph{Actions}

\begin{figure}[ht]
  \centering
  \includegraphics[width=0.75\textwidth]{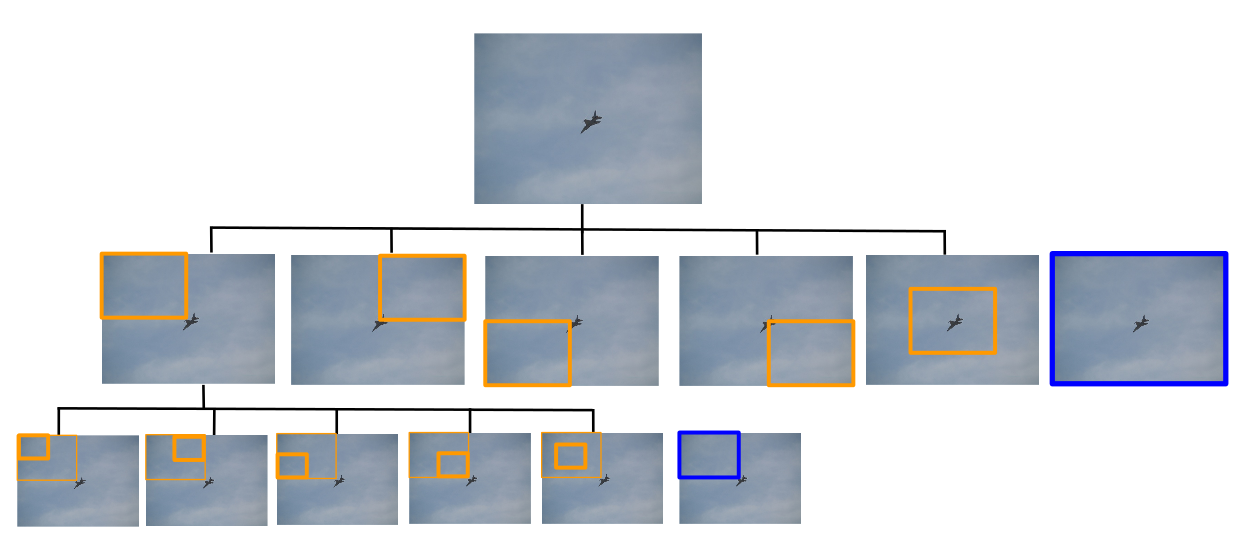}
  \caption{Image hierarchy of three levels with non-overlapped quarters}
  \label{part1}
\end{figure}

\begin{figure}[ht]
  \centering
  \includegraphics[width=0.75\textwidth]{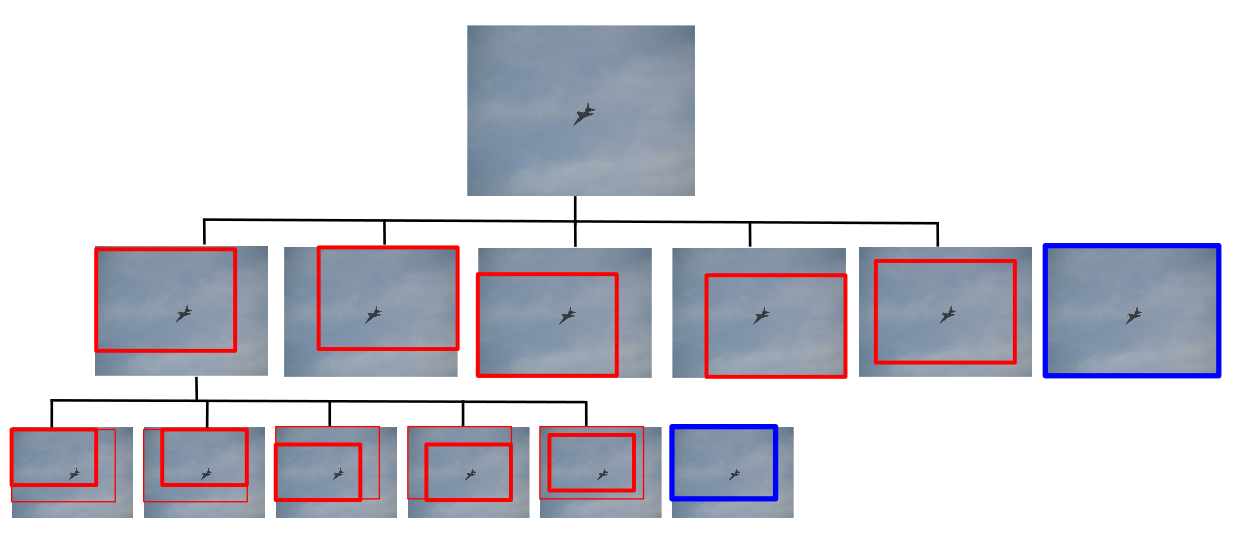}
  \caption{Image hierarchy of three levels with overlapped quarters}
  \label{part2}
\end{figure}

There are two types of possible actions: \textit{movement actions} that imply a change in the current observed region, and the \textit{terminal} action to indicate that the object is found and that the search has ended. 
One particularity of our system is that each movement action can only transfer the attention top-down between regions from a predefined hierarchy. A hierarchy is built by defining five subregions over each observed bounding box: four quarters distributed as $2x2$ over the box and a central overlapping region.
We have explored two variations of this basic $2x2$ scheme: a first one with \textit{non-overlapped} quarters (see Figure ~\ref{part1}), and a second one with \textit{overlapped} quarters (see Figure ~\ref{part2}), being the size of a subregion 3/4 of its ancestor.
Then, there are five movement actions, each one associated to one of the yellow regions. If, on the other hand, the terminal action is selected, there is no movement and the final region is the one marked with blue. 

\paragraph{Reward}

The reward functions used are the ones proposed by  \citet{caicedo2015active}. The reward function for the movement actions can be seen in Equation \ref{eq:reward1} and the reward function for the terminal action in Equation \ref{eq:reward2}. Given a certain state \textit{s}, a reward is given to those actions that move towards a region \textit{b'} with a greater Intersection Over Union (IoU) with the ground truth \textit{g} than the region \textit{b} considered at the previous step. Otherwise, the actions are penalized. For the trigger action, the reward is positive if the Intersection Over Union of the actual region \textit{b} with the ground truth is greater than a certain threshold \(\tau\), and negative otherwise. We consider \(\tau\) = 0.5 , because it is the threshold for which a detection is considered positive, and \(\eta\) is 3, as in \citep{caicedo2015active}.

\begin{equation} \label{eq:reward1}
R_m(s,s') = sign(IoU(b',g)-IoU(b,g))
 \end{equation}
 
\begin{equation} \label{eq:reward2}
 R_t(s,s') =\left\{
                \begin{array}{ll}
                  +\eta  \hspace{0.7cm} if \hspace{0.1cm} IoU(b,g) \geq \tau \\
                  -\eta  \hspace{0.7cm}  otherwise \\
                \end{array}
              \right.
 \end{equation}

\subsection{Q-learning}

The reward of the agent depending on the chosen action \textit{a} at state \textit{s} is governed by a function \textit{Q(s,a)}, that can be estimated with Q-learning.
Based on \textit{Q(s,a)}, the agent will choose the action that is associated to the highest reward. Q-learning iteratively updates the action-selection policy using the Bellman equation \ref{eq:bellman}, where \textit{s} and \textit{a} are the current state and action correspondingly, \textit{r} is the immediate reward and \(max_{a'} Q(s', a')\) represents the future reward. Finally \(\gamma\) represents the discount factor. In our work, we approximate the Q-function by a Deep Q-network trained with Reinforcement Learning \citep{mnih2015human}.

\begin{equation} \label{eq:bellman}
Q(s,a) = r + \gamma max_{a'} Q(s', a')
 \end{equation}

\subsection{Model}
\label{model_section}

\begin{figure}[ht]
  \centering
  \includegraphics[width=0.80\textwidth]{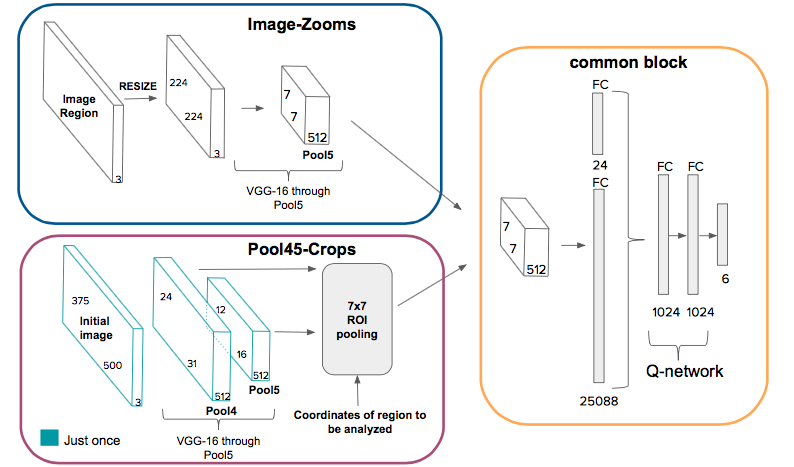}
  \caption{Hierarchical Object Detection Models}
  \label{allmodel}
\end{figure}

In our work we study two different approaches for visual feature extraction, which are used to train a Deep Q-Network. Figure ~\ref{allmodel} depicts the two variations with the common reinforcement learning network.

We compare two models to extract the visual features that define the state of our agent: the \textit{Image-Zooms} model and the \textit{Pool45-Crops} model.
For the \textit{Image-Zooms} model, each region is resized to 224x224 and its visual descriptors correspond to the feature maps from Pool5 layer of VGG-16 \citep{Simonyan14c}. For the \textit{Pool45-Crops} model, the image at full-resolution is forwarded into VGG-16 \citep{Simonyan14c} through Pool5 layer. As \citet{girshick2015fast}, we reuse the feature maps extracted from the whole image for all the  regions of interest (ROI) by pooling them (ROI pooling). 
As in SSD \citep{liu2015ssd}, we choose which feature map to use depending on the scale of the region of interest. In our case, we only work with the Pool4 and Pool5 layers, that are the two last pooling layers from VGG-16. Once we have a certain region of interest from the hierarchy, we decide which feature map to use by comparing the scale of the ROI and the scale of the feature map. For large objects, the algorithm will select the deeper feature map, whereas for smaller objects a shallower feature map is more adequate. 

The two models for feature extraction result into a feature map of 7x7 which is fed to the common block of the architecture. 
The region descriptor and the memory vector are the input of the Deep Q-network that consists of two fully connected layers of 1024 neurons each. Each fully connected layer is followed by a ReLU \citep{nair2010rectified} activation function and is trained with dropout \citep{srivastava2014dropout}. Finally the output layer corresponds to the possible actions of the agent, six in our case.

\subsection{Training}

In this section we will explain the particularities that we chose to train the Q-network.

\paragraph{Exploration-Exploitation} To train the deep Q-network with reinforcement learning we use an \(\epsilon\)-greedy policy, that starts with \(\epsilon\)=1 and decreases until \(\epsilon\)=0.1 in steps of 0.1. Then, we start with random actions, and at each epoch the agent takes decisions relying more on the already learnt policy. Actually, in order to help the agent to learn the terminal action, which in random could be difficult to learn, we force it each time the current region has a IoU > 0.5. With this approach we can accelerate the training. Notice that we always do exploration, so we do not get stuck into a local minimum. 

\paragraph{Learning trajectories} One fact that we detected while training was that we should not impose which object of the image to look first.
At each time step, the agent will focus on the object in the current region with the highest overlap with its ground-truth. This way, it is possible then that the target object changes during the top-down exploration.

\paragraph{Training parameters} The weights for the Deep Q-network were initialized from a normal distribution. For learning, we used Adam optimizer \citep{kingma2014adam} with a learning rate of 1e-6 to avoid that the gradients explode. We trained each model for 50 epochs.

\paragraph{Experience Replay} As we have seen previously, Bellman Equation \ref{eq:bellman} learns from transitions formed by \textit{(s, a, r, s')}, which can also be called experiences. Consecutive experiences in our algorithm are very correlated and this could lead to inefficient and unstable learning, a traditional problem in Q-learning. One solution to make the algorithm converge is collecting experiences and storing them in a replay memory. Random minibatches from this replay memory are used to train the network. We used an experience replay of 1,000 experiences and a batch size of 100.

\paragraph{Discount factor} To perform well in the long-run, the future rewards should also be taken into account and not only the most immediate ones. To do this, we use the discounted reward from Bellman Equation \ref{eq:bellman} with a value of \(\gamma\) = 0.90. We set the gamma high because we are interested in balancing the immediate and future rewards.

\section{Experiments}
\label{experiment}

Our experiments on object detection have used images and annotations from the PASCAL VOC dataset \citep{everingham2010pascal}. 
We trained our system on the trainval sets of 2007 and 2012, and tested it on the test set of 2007. We performed all the experiments for just one class, the \textit{aeroplane} category, and only considering pictures with the target class category.
This experiment allows us to study the behavior of our agent and estimate the amount of regions that must be analyzed to detect an object.

\subsection{Qualitative results}

\begin{figure}[ht]
  \centering
  \includegraphics[width=0.75\textwidth]{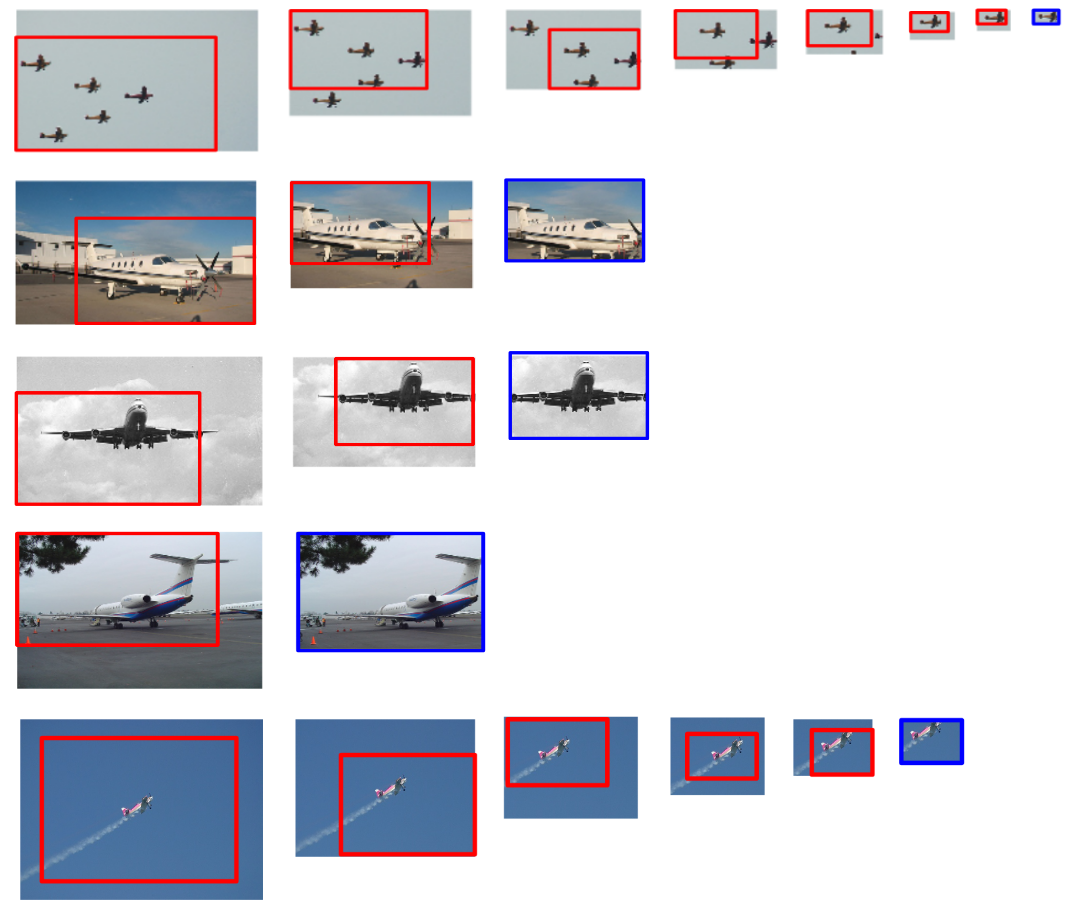}
  \caption{Visualizations of searches for objects }
  \label{fig:visualization}
\end{figure}

We present some qualitative results in Figure ~\ref{fig:visualization} to show how our agent behaves on test images.
These results are obtained with the \textit{Image-Zooms} model with overlapped regions, as this is the one that yields best results, as argued in the following sections. 
We observed that for most images, the model successfully zooms towards the object and completes the task in a few steps. As seen in the second, third and fourth rows, with just two or three steps, the agent selects the bounding box around the object. The agent also performs accurately when there are small instances of objects, as seen in the first and last rows.

\subsection{Precision-Recall curves}

We will analyze the precision and recall curves for different trained models, considering that an object is correctly detected when the Intersection over Union (IoU) of its bounding box compared to the ground truth is over 0.5, as defined by the Pascal VOC challenge \citep{everingham2010pascal}. 

The Precision-Recall curves are generated by ranking all regions analyzed by the agent. The sorting is based on the reward estimated by the sixth neuron of the Q-network, which corresponds to the action of considering the region as terminal.

\begin{figure}[!tbp]
  \centering
  \begin{minipage}[b]{0.45\textwidth}
    \includegraphics[width=\textwidth]{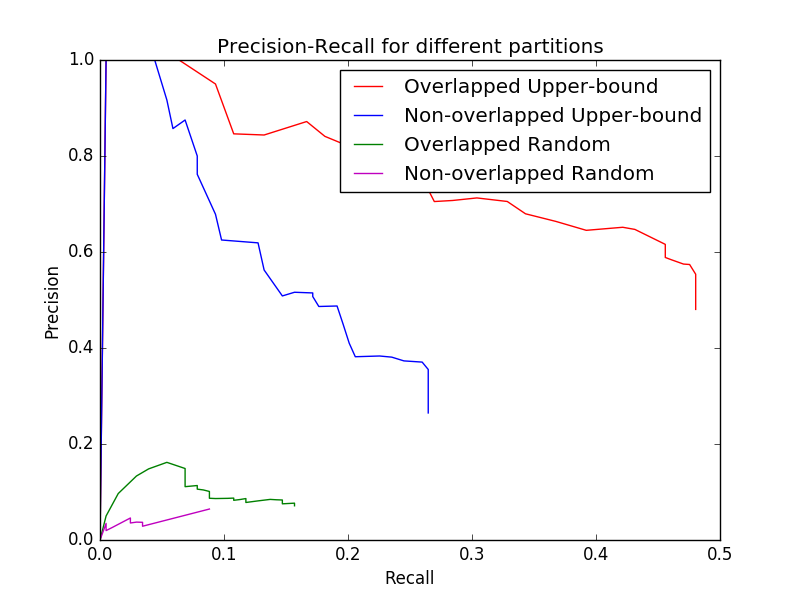}
    \caption{Baselines and upper-bounds}
    \label{baselines}
  \end{minipage}
  \hfill
  \begin{minipage}[b]{0.45\textwidth}
    \includegraphics[width=\textwidth]{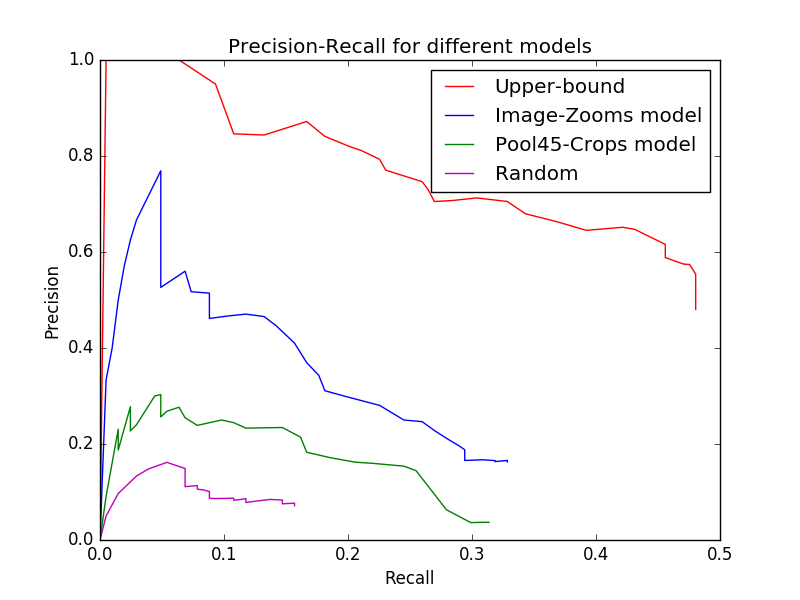}
    \caption{Comparison of the two models}
    \label{models_compare}
  \end{minipage}
\end{figure}

\paragraph{Upper bound and random baselines}
Our results firstly include baseline and upper-bound references for a better analysis.
As a baseline we have programmed an agent that chooses random actions and detection scores at each time step. As an upper-bound, we have exploited the ground truth annotations to manually guide our agent towards the region with the greatest IoU.
The result of these random baselines and upper-bounds for hierarchy type can be seen in Figure ~\ref{baselines}. 
It is also important to notice that the best upper-bound option does not even achieve a a recall of 0.5. This poor performance is because more than half of the ground truth objects do not fit with the considered region proposals, so they cannot be detected in our framework. 


\paragraph{Overlapped and non-overlapped regions}
The results obtained with the upper-bound and baseline methods provide enough information to compare the overlapped and non-overlapped schemes.
The overlapped regions scheme is the one that provides higher precision and recall values, both for the upper-bound and the random models.
This superiority of the overlapped case can be explained by the slower reduction of spatial scale with respect to the non-overlapped model: as bounding box regions are larger due to the overlap, their division in equal-size subregions also generate larger subregions. This also implies that the agent will require more steps to reach a lower resolution, but this finer top-down exploration is shown as beneficial in our experiments as the chances of missing an object during the descent are also lower.



\paragraph{Model comparison}

The \textit{Image-Zooms} model and the \textit{Pool45-Crops} model are compared in Figure ~\ref{models_compare}.
Results clearly indicate that the \textit{Image-Zooms} model performs better than the \textit{Pool45-Crops} model. 
We hypothesize that this loss of performance is due to the loss of resolution resulting from the ROI-pooling over Pool4 or Pool5 layers.
While in the \textit{Image-Zooms} model the 7x7 feature maps of Pool5 have been computed directly from a zoom over the image, in the \textit{Pool45-Crops} model the region crops over Pool4 or Pool5 could be smaller than 7x7.
While these cases would be upsampled to the 7x7x512 input tensor to the deep Q-Net, the source feature maps of the region would be of lower resolution than their counterparts in the \textit{Image-Zoom} model.


\paragraph{Models at different epochs}

\begin{figure}[!tbp]
  \centering
  \begin{minipage}[b]{0.45\textwidth}
    \includegraphics[width=\textwidth]{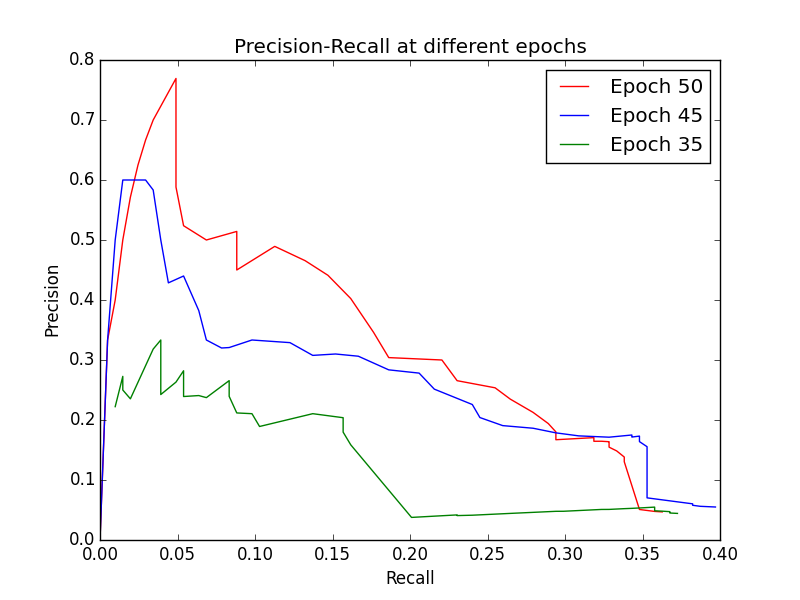}
    \caption{Image-Zooms at different epochs}
    \label{different_epochs}
  \end{minipage}
  \hfill
  \begin{minipage}[b]{0.45\textwidth}
    \includegraphics[width=\textwidth]{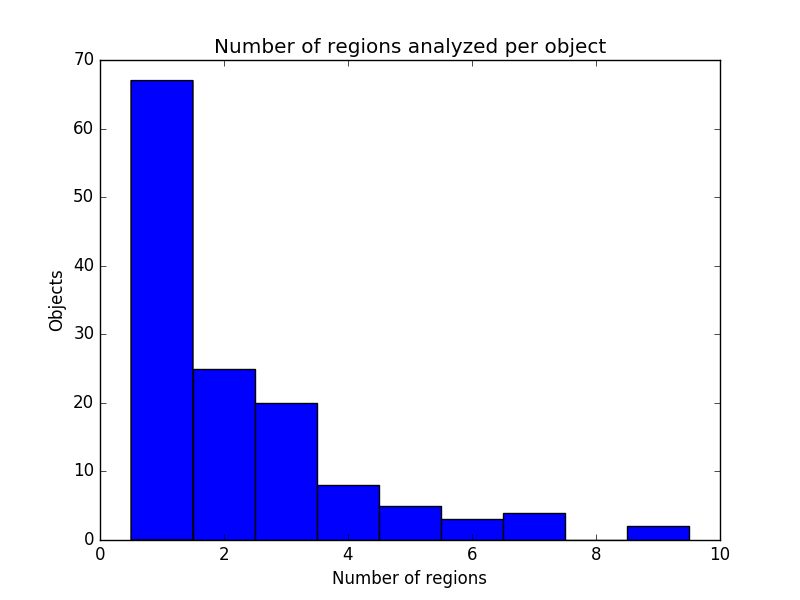}
    \caption{Regions analyzed per object}
    \label{fig:histogram}
  \end{minipage}
\end{figure}

We study the training of our \textit{Image-Zooms} model by plotting the Precision-Recall curves at different epochs in Figure ~\ref{different_epochs}. 
As expected, we observe how the performance of the model improves with the epochs.


\subsection{Number of regions analyzed per object}

An histogram of the amount of regions analyzed by our agent is shown in ~\ref{fig:histogram}.
We observe that the major part of objects are already found with a single step, which means that the object occupies the major part of the image. With less than 3 steps we can almost approximate all objects we can detect.

\section{Conclusions}
\label{discussion}

This paper has presented a deep reinforcement learning solution for object detection.
Our solution is characterized by a top-down exploration of a hierarchy of regions guided by an intelligent agent.

Our experiments indicate that objects can be detected with very few proposals from an appropriate hierarchy, but that working with a predefined set of regions clearly limits the recall. A possible solution to this problem would be refining the approximate detections provided by the agent with a regressor, as in \citep{ren2015faster}.

Finally, our results indicate the limitations of cropping region features from the convolutional layers, especially when considering small objects. We suggest that, given the much smaller amount of region proposals considered by our reinforcement learning agent, feeding each region through the network is a solution that should be also considered. 

The presented work is publicly available for reproducibility and extension at \url{https://imatge-upc.github.io/detection-2016-nipsws/}.


\section*{Acknowledgments}
\label{acknowledgements}

This work has been developed in the framework of the project BigGraph TEC2013-43935-R, funded by the Spanish Ministerio de Economia y Competitividad and the European Regional Development Fund (ERDF). This work has been supported by the grant SEV2015-0493 of the Severo Ochoa Program awarded by Spanish Government, project TIN2015-65316 by the Spanish Ministry of Science and Innovation contracts 2014-SGR-1051 by Generalitat de Catalunya. The Image Processing Group at the UPC is a SGR14 Consolidated Research Group recognized and sponsored by the Catalan Government (Generalitat de Catalunya) through its AGAUR office. We gratefully acknowledge the support of NVIDIA Corporation with the donation of the GeForce GTX Titan Z used in this work and  the support of BSC/UPC NVIDIA GPU Center of Excellence. We also want to thank all the members of the X-theses group for their advice.

\small

\bibliographystyle{plainnat}
\bibliography{ex}

\end{document}